\documentclass[pra,aps,superscriptaddress,nofootinbib,twocolumn]{revtex4-1}

\usepackage{mathrsfs}
\usepackage{amsfonts}
\usepackage{amssymb}
\usepackage{amsmath}
\usepackage{graphicx}
\usepackage[usenames,dvipsnames]{color}
\usepackage{ulem}
\usepackage{lmodern}
\usepackage{amsthm}
\usepackage{xcolor}
\usepackage{subfigure}
\usepackage{url}
\usepackage{hyperref}

\usepackage{tikz}

\usetikzlibrary{chains}
\hypersetup{
    colorlinks=true,
    linkcolor=blue,
    filecolor=blue,      
    urlcolor=blue,
    citecolor=cyan,
}

\newcommand{\norm}[1]{\Vert #1 \Vert}

\newcommand{\ket}[1]{\vert{ #1 }\rangle}
\newcommand{\bra}[1]{\langle{ #1 }\vert}

\newtheorem{theorem}{Theorem}

\begin{document}
%\tikzset{mark size=0.1}
\title{unsupervised machine learning for physical concepts}

\author{Ruyu Yang}
\affiliation{Graduate School of China Academy of Engineering Physics, Beijing 100193, China}

%\captionsetup[figure]{name={Fig.}} 

\begin{abstract}
In recent years, machine learning methods have been used to assist scientists in scientific research. Human scientific theories are based on a series of concepts. How machine learns the concepts from experimental data will be an important first step. We propose a hybrid method to extract interpretable physical concepts through unsupervised machine learning. This method consists of two stages. At first, we need to find the Betti numbers of experimental data. Secondly, given the Betti numbers, we use a variational autoencoder network to extract meaningful physical variables. We test our protocol on toy models and show how it works. 
\end{abstract}

\maketitle

\section{introduction}

Physics is built on many different concepts, such as force, entropy, and Hamiltonian. Scientists derive meaningful concepts from observations by their ingenuity and then use formulas to connect them, constituting the theory. As a principle, this theory is always as simple as possible. Since the beginning of the 21st century, machine learning has developed rapidly~\cite{liu2018artificial} and has been widely used in various fields~\cite{teichert2019machine,bourilkov2019machine,mishra2016applications}, including machine-assisted scientific research~\cite{vamathevan2019applications,valletta2017applications,schmidt2019recent,carleo2019machine}. A natural question arises: can machines propose scientific theories by themselves? It is undoubtedly a fundamental and challenging goal. Many works have studied this from different aspects~\cite{ha2021discovering,zhang2018robust,boulle2021data,zobeiry2019iterative,farina2020searching,wu2019toward,zheng2018unsupervised,d2019learning}. In general, the establishment of the theory can be divided into two steps: identify the critical variables from observations and connect them by formula. A technique known as symbolic regression has been developed for the second step~\cite{udrescu2020ai01,udrescu2020ai02}. The authors propose a network named AI Feymann to automatically find the formula the system obeys. To increase the success rate of symbolic regression, the critical variables identified by the first step should be as few as possible.

%So we always hope the concepts are as few as possible. 

%With the development of machine learning, it is possible to use machines to help with discovering physical concepts using experimental observations. To achieve this aim, some approaches have been developed. A method is to use variational autoencoder(VAE)  neuron network to extract physical concepts\cite{}.

In this work, we focus on the first step of establishing a theory. We suggest using both Topological Data Analysis(TDA)\cite{wasserman2018topological,chazal2017introduction,murugan2019introduction} and variational autoencoder(VAE) \cite{kingma2019introduction,kingma2013auto,an2015variational,khobahi2019model,rezende2014stochastic,higgins2016beta,burgess2018understanding} to extract meaningful variables from extensive observations. TDA is a recent and fast-growing tool to analyze and exploit the complex topological and geometric structures underlying data. This tool is necessary for our protocol if specific structures such as circles and sphere are in experimental data. VAE is a deep learning technique for learning latent representations. This technique has been widely used in many problems\cite{luchnikov2019variational,cerri2019variational} as a generative model. It can also be seen as a manifold learning network for dimensionality reduction and unsupervised learning tasks.

Our protocol has two stages. Firstly, we use TDA to infer relevant topological features for experimental data. In the simplest case, where the manifold has a low dimension, an essential feature for us is the Betti numbers, topology invariants. Once we get the topological features, we can design the proper architecture and loss function of VAE. As we will show later, the latent variables of VAE need to form a manifold homeomorphic to the manifold composed of observations. After the training of the VAE network, the latent variables represent the key variables discovered by this machine. Thanks to the structure of the VAE network, the latent variables, and the observations are in one-to-one correspondence. Using the trained VAE network, one can calculate the latent variables and the observations from each other. That means the formula derived by symbolic regression connecting the latent variables can predict the experimental phenomena.

We test our protocol on three toy models. They have different topological features. The first is a classical coupled harmonic oscillator, where the observations constitute a circle embedded in three-dimensional Euclidean space. Another example is two balls rotating around the same center, with different radius and angular velocities. With the other ball as the reference system, the observations are the Cartesian coordinates of the ball. The observations constitute a lemniscate curve. The third is a two-level system, and the observations are the expected value of some physical quantity,i.e., some hermitian matrices. The observations constitute a sphere.
%Except for the value of latent variants, we can also derive the distribution, which can be the foundation for discovering the formula.

%We focus our attention on discovering the concepts in Quantum Mechanics. We consider three toy models to test our protocol. As a result, the machine will find the Hamiltonian and predict the spectral finally.

This paper is organized as follows. In section \ref{goal}, we describe the works been done before and the problem we want to solve in more detail.
In section \ref{vae}, we introduce the architecture of neuron networks and argue why we need the manifold of latent variants should be homeomorphic to that of observations. In section \ref{numerical}, we show the performance of this protocol on three toy models. We compare the observations and latent variables to show the relation between them.%To demonstrate that the latent variants are meaningful, we use the latent variants of the first model as the observations of the second model, and so on.

\section{Related work and our goal}
\label{goal}
Data-driven scientific discoveries are not new ideas. It follows from the revolutionary work of Johannes Kepler and Sir Isaac Newton~\cite{roscher2020explainable}. Unlike the 17th century, we now have higher quality data and better calculation tools. People have done a lot of research on how to make machines help people make scientific discoveries in different contexts~\cite{carleo2019machine,cichos2020machine,karpatne2018machine,wetzel2017unsupervised,rodriguez2019identifying,wu2019toward}. In the early days, people paid more attention to the symbolic regression~\cite{kim2020integration,udrescu2020ai01,udrescu2020ai02,lample2019deep}. Another challenging direction is to let the machine design experiments. In ~\cite{melnikov2018active,krenn2016automated,fosel2018reinforcement,bukov2018reinforcement}, authors designed automated search techniques and a reinforcement-learning-based network to generate new experimental setups. In condensed physics, machine learning has been used to characterize phase transitions~\cite{van2017learning,uvarov2020machine,carrasquilla2017machine,wang2016discovering,ch2017machine,yu2021unsupervised}.

In recent years, some works have contributed to letting the machine search for key physical concepts. They used the different networks to extract key physical parameters from the experimental data~\cite{iten2020discovering,zheng2018unsupervised,lu2020extracting}.
In \cite{zheng2018unsupervised}, authors propose an unsupervised method to extract physical parameters by interaction networks~\cite{battaglia2016interaction}. Another helpful tool is the VAE network, which has been widely used for similar goal~\cite{lu2020extracting,iten2020discovering}.

VAE network is a powerful tool, by which one can minimize the numbers of extracted variables by making the variables independent of each other ~\cite{higgins2016beta,burgess2018understanding}. To do this, one can choose the prior distribution $P(z)$ as latent variables $z$. However, this method fails when the manifold of observations have some topological features. This paper aims to solve this problem.
%~\cite{} propose a unsupervised method to extract physical parameters from the system obeying classical statistical mechanics.

Here we describe this goal in more detail. Suppose we have an experimental system $\mathcal{S}$. In this system $\mathcal{S}$, some physical variables change as the experiment progresses. We use $\mathcal{P}$ to denote the set consists of the value of these physical variables. We use $\mathcal{P}^{(k)}$ to represent the value of these variables in time $k$.
From the system $\mathcal{S}$ we can derive an experimental data set $\mathcal{E}$. Every data point is a vector $\mathcal{E}^{(k)} \in \mathcal{E}$ where $\mathcal{E}^{(k)}$ denotes the data point belongs to time $k$. From the perspective of physics, the change of experimental data must be attributed to the change of key physical variables. Therefore there is a function $f: \mathcal{P} \rightarrow \mathcal{E}$ such that $f(\mathcal{E}^{(k)}) = \mathcal{P}^{(k)}$ for any $k$. In this paper we aim to find the function $\Tilde{f}$ and $\Tilde{f}^{-1}$ such that $\Tilde{\mathcal{P}}^{(k)} = \Tilde{f}^{-1} (\mathcal{S}^{(k)})$ where $\Tilde{\mathcal{P}}^{(k)}$ constitute a set $\Tilde{\mathcal{P}}$. We call $\Tilde{\mathcal{P}}^{(k)}$ the effective physical variables. We remark that $\Tilde{\mathcal{P}}^{(k)}$ is not necessary to equal to $\mathcal{P}^{(k)}$, but their dimensions should be the same. Effective physical variables are enough to describe the experimental system. In fact, one can redefine the physical quantity can get a theory that looks very different but the predicted results are completely consistent with existing theories.

The problem arise if we try to use neuron network to find proper function $\Tilde{f}$ and $\Tilde{f}^{-1}$. All the functions neuron network can simulate is continuous\ref{}, so $\Tilde{f}$ and $\Tilde{f}^{-1}$ must be continuous. In a real physical system, $f$ and $f^{-1}$ don't have to be like this. Therefore, in some cases, we can never find a $\Tilde{\mathcal{P}}^{(k)}$ which has the same dimension with $\mathcal{P}^{(k)}$. For example, suppose we have a ball rotating around a center. The observable data is the location of the ball, denoted by Cartesian coordinates $\mathcal{E}^{(k)} = (x^{(k)},y^{(k)})$, which form a circle, and the simplest physical variable is the angle $\mathcal{P}^{(k)} = \theta^{(k)} \in [0,2\pi)$. In this case, $f: \theta \rightarrow (x,y)$ is continuous while $f^{-1}$ is not, which cannot be approximated by neural network. This is because $f$ is a periodic function of $\theta$. More generally, as long as the manifold composed of $\mathcal{E}$ has holes with any dimension, this problem arises. We call these physical variables topological physical variables(TPVs). Back to the last example, we can find the Betti numbers of the ball's location is $[1,1,0]$, where the second $1$ means that it has one TPVs. Due to this reason, we suggest using TDA to identify the topological features firstly. After knowing the numbers of TPVs, we can design proper latent variables and the corresponding loss function $\mathcal{L}$. For the case we have two PPVs, we need two latent variables, named $x$ and $y$, and we add the topological term $|x^2 + y^2 - 1|$ to $\mathcal{L}$ to restrict them.

In at least two cases, the manifold will have holes. One is the $\mathcal{E}$ of a conversed system,e.g., the classical harmonic oscillator. Another situation stems from the limits of the physical quantity itself, such as the single-qubit state, which forms a sphere. In addition to these two categories, sometimes the choice of observations and references also affects their topological properties.

%PPVs are widely found in nature. When there are enough observations, PPVs are always associated with the topology.
%Back to the last example, we can find the Betti numbers of the location of the ball is $[1,1,0]$, where the second $1$ means that it has one PPV. Due to this reason, we suggest to firstly use TDA to identify the topological features. After knowing the numbers of PPVs, we can design proper latent variables and the corresponding loss function $\mathcal{L}$. For the case we have one PPV, we need two latent variables, named $x$ and $y$, and we add the topological term $|x^2 + y^2 - 1|$ to $\mathcal{L}$ to restrict them.

%Several work have been developed for extracting physical variables from data set. Some work focus on the mathematical expression corresponding to given $\mathcal{P}$ and $\mathcal{E}$\cite{}. A series of technology called symbolic regression has been developed for this goal\cite{}. On the other hand, some papers focus on letting machine think like human \cite{}. In \cite{}, the authors suggest to use VAE to extract efficient physical variables. However, traditional VAE network can only look for physical variables that have no relation to each other. But in some cases, like our examples, due to the limitations of neuron network, VAE can never find such physical variables. That's why we need to analyze the topological features of observations.h

%\includegraphics[scale=1]{output-figure14.pdf}
\section{architecture of neuron network}
\label{vae}
 
\begin{figure}[tbp]
\centering
    \begin{tikzpicture}
    \draw [->] (0,0,0) --(0,0,1);
    \draw [->] (0,0,0) --(0,1,0);
    \draw [->] (0,0,0) --(1,0,0);
    \draw [-][draw = red] (0.5,0.5,0.5) to [out=30, in = 150] (2,3,4) ;
    \draw [-][draw = red] (2,3,4) to [out=-60, in = 100] (3,2,3) ;
    \draw [-][draw = red] (3,2,3) to [out=-80, in = 0] (2,1,3) ;
    \draw [-][draw = red] (2,1,3) to [out=180, in = 210] (0.5,0.5,0.5) ;
    \draw [->][blue,ultra thick] (2,0.5) --(3,0.5);
    \draw [->] (4,0.5) --(4,1.5);
    \draw [->] (4,0.5) --(5,0.5);
    \draw [red] (4,0.5) circle [radius = 0.5];
    \node at (1,2) {observations};
    \node at (4,2) {latent vaxriables};
    \node at (2.5,0) {Encoder};
    \end{tikzpicture}
    
    \caption{
The observations of a system form a form a closed curve in a three-dimensional space. We can encoder it as a two-dimensional circle. However, we don’t have to encode it as a circle. Ellipse or other homeomorphic curves are all allowed.}
    \label{circle}
\end{figure}
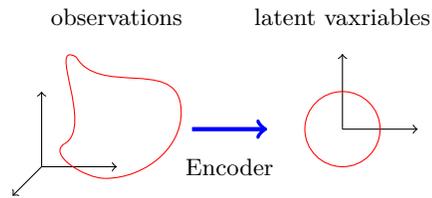
\begin{figure}
    \begin{tikzpicture}
    \draw [->] (0,0) --(1,0);
    \draw [->] (0,1) --(1,1);
    \draw [->] (0,2) --(1,2);
    \draw [->] (0,3) --(1,3);
    \draw [blue] (1.3,0) circle [radius = 0.3];
    \draw [blue] (1.3,1) circle [radius = 0.3];
    \draw [blue] (1.3,2) circle [radius = 0.3];
    \draw [blue] (1.3,3) circle [radius = 0.3];
    \draw [red] (2.7,1) circle [radius = 0.3];
    \draw [red] (2.7,2) circle [radius = 0.3];
    \draw [->] (1.6,0) --(2.4,1);
    \draw [->] (1.6,1) --(2.4,1);
    \draw [->] (1.6,2) --(2.4,1);
    \draw [->] (1.6,3) --(2.4,1);
    \draw [->] (1.6,0) --(2.4,2);
    \draw [->] (1.6,1) --(2.4,2);
    \draw [->] (1.6,2) --(2.4,2);
    \draw [->] (1.6,3) --(2.4,2);
    \draw [blue] (4,0) circle [radius = 0.3];
    \draw [blue] (4,1) circle [radius = 0.3];
    \draw [blue] (4,2) circle [radius = 0.3];
    \draw [blue] (4,3) circle [radius = 0.3];
    \draw [->] (3,1) --(3.7,0);
    \draw [->] (3,1) --(3.7,1);
    \draw [->] (3,1) --(3.7,2);
    \draw [->] (3,1) --(3.7,3);
    \draw [->] (3,2) --(3.7,0);
    \draw [->] (3,2) --(3.7,1);
    \draw [->] (3,2) --(3.7,2);
    \draw [->] (3,2) --(3.7,3);
    \draw [->] (4.3,3) --(5,3);
    \draw [->] (4.3,2) --(5,2);
    \draw [->] (4.3,1) --(5,1);
    \draw [->] (4.3,0) --(5,0);
    \node at (-0.5,1.5) {input $x$};
    \node at (5.5,1.5) {output $\hat{x}$}; 
    \node at (1.3,-0.5) {Encoder};
    \node at (4,-0.5) {Decoder};
    \node at (2.7,0.5) {Latent};
    \end{tikzpicture}
    \caption{There is a bottleneck in the network which forces a compressed knowledge representation. One needn't to set the decoder part as the inverse of encoder part, and the networks of encoder and decoder don't have to be the same. In general, the dimension of output $\hat{x}$ should be the same as the input $x$. The number of hidden variables is usually less than the input.}
    \label{network}
\end{figure}
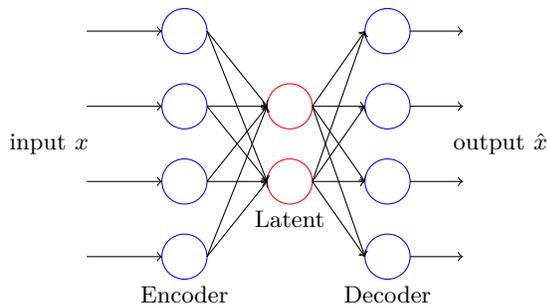

As shown in figure ~\ref{network}, This network consists of two parts. The left is called encoder, and the right is called decoder. The encoder part is used to encode high-dimensional data into low-dimensional space, while the decoder part is used to recover the data, i.e., the input from low-dimensional data. The latent variables are low-dimensional data, and we hope they contain all the information needed to recover the input data. So that we require the output $\hat{x}$ is as close as possible to input $x$. Therefore, we can train this neuron network without supervision.

In general, the encoder and decoder can simulate arbitrary continuous function $f$ and $f^{-1}$. That imposes restrictions on latent variables: 
\begin{theorem}
The manifold composed of latent variables must be homeomorphic to that of input $x$.
\label{homeo}
\end{theorem}

This limitation means that the Betti numbers should keep the same. Suppose the inputs $x$ form a circle, as shown in \ref{circle}, the easiest quantity to describe these data is the angle. According to \ref{homeo}, this neuron network can never find such a quantity because the angle will form a line segment that is not homeomorphic to a circle. On the other hand, VAE network can't even find the Cartesian coordinate $\{x_1,x_2\}$ because $x_1$ and $x_2$ are not irrelevant.  To handle this case, we suggest analyzing the Betti numbers as a priori knowledge of input $x$ and tell the neuron network.%, as shown in figure ~\ref{}.

The general loss function is wirtten as~\cite{higgins2016beta}:
\begin{equation}
\mathcal{L}=-\alpha \mathbb{E}_{q_{\phi}(\mathbf{z} \mid \mathbf{x})}\left[\log p_{\theta}(\mathbf{x} \mid \mathbf{z})\right]+\beta D_{K L}\left(q_{\phi}(\mathbf{z} \mid \mathbf{x}) \| p(\mathbf{z})\right)
\end{equation}
Here $\theta,\phi$ are the parameters of decoder and encoder networks, respectively. $p(z)$ is the prior distribution about the latent variables $z$. $q_{\phi} (z|x)$ and $p_{\theta}(x|z)$ denote the conditional probability. $D_{K L}$ means the KL divergence of two distributions. 
\begin{equation}
    D_{K L}\left(q_{\phi}(\mathbf{z} \mid \mathbf{x}) \| p(\mathbf{z})\right) = \mathbb{E}_{q_{\phi}(\mathbf{z} \mid \mathbf{x})}(\log q_{\phi}(\mathbf{z} \mid \mathbf{x}) - \log p(\mathbf{z})) %= H(q,p) - H(q)
\end{equation}

% In traditional VAE, to minimize the number of latent variables, the prior $P(z)$ is set to be independent Gaussian distribution. We need to change the $P(z)$ according to the topological features to handle the case we consider. We do this by classifying the latent variables. 
To calculate the loss function for a given output $\hat{x}$, one needs to set up a prior $p(z)$, and parameterize the distribution $p_{\theta}$ and $q_{\phi}$. In traditional VAE, the distributions $p(z)$, $p_{\theta}$ and $q_{\phi}$ are often supposed to be multidimensional Gaussian distribution. The mean of $p(z)$ is zero the and variance is $1$. The mean and variance of $p_{\theta}$ and $q_{\phi}$ is the output of encoder network. 

In our cases, we choose the prior $p(\Vec{z})$ as $p(\Vec{z}) = p(\Vec{z_g}) \times p(\Vec{z_t})$. Where $\Vec{z_g}$ denotes the GPVs, and $\Vec{z_t}$ denotes the TPVs. With the same as traditional VAE, in order to minimize the number of GPVs, a convenient way is to choose $p(\Vec{z_g})$ as Gaussion distribution $p(\Vec{z_g}) = N(0,I)$. According to the topological features, we choose a proper topological term $\mathcal{T}$. We choose the $p(\Vec{z_t})$ as $p(\Vec{z_t}) = A e^{-T}$. Here $A$ is a constant. The variance can be asorpted into the hyperparameters in loss function.

For the conditional distribution we choose as s $q_{\phi}(\mathbf{z} \mid \mathbf{x}) = N(\Vec{z} , I) $, with constantly variance and means determined by the encoder network. $p_{\theta}(\mathbf{x} \mid \mathbf{z})$ has the same expression $p_{\theta}(\mathbf{x} \mid \mathbf{z}) = N(x,I)$.
In this assumption, the KL divergence can be written as 
\begin{equation}
    D_{K L}\left(q_{\phi}(\mathbf{z} \mid \mathbf{x}) \| p(\mathbf{z})\right) = - \mathbb{E}_{q_{\phi}(\mathbf{z} \mid \mathbf{x})}(\log p(\mathbf{z})) + constant
\end{equation}

Thus the loss function can be written as a simplier expression:
\begin{equation}
    \mathcal{L} = \alpha \norm{x - \hat{x}}_2 + \beta \mathcal{T}  - \gamma log(P(\Vec{z_g}))
    \label{loss}
\end{equation}
where $\alpha,\beta,\gamma$ are hyperparameters and $\mathcal{T}$ is the topological term depending on the Betti numbers. For example, when the Betti numbers are $\{1,0,1\}$, $\mathcal{T}$ can be written as $\mathcal{T} = |z_1^2 + z_2^2 + z_3^2 - 1|$. When the Betti numbers are $\{1,2,0\}$, $\mathcal{T}$ can be written as $\mathcal{T} = |(z_1^2 + z_2^2) - 2 (z_1^2 -z_2^2)|$, which is known as a Lemniscate. Here we use $\Vec{z}$ to denote the latent variables. $P(\Vec{z_g})$ is the distribution of $\Vec{z_g}$. We need to choose some latent variables as TPVs, and others as general physical variables(GPVs). In general, we know how many TPVs we need from the Betti numbers, but we usually don't know how many GPVs we need. One solution is to set up as many GPVs as possible, and the redundant GPVs will be zero. 
\section{Numerical Simulation}
\label{numerical}
\begin{figure*}[ht!]
%   \centering
%   \parbox{5in}{%
% \includegraphics[width=0.2\linewidth]{xiezhenzi_latent.pdf}
% \caption{First.}%
% \label{fig:2figsA}}%
% \qquad
% \begin{minipage}{5in}%
% \includegraphics[width=0.2\linewidth]{xiezhenzi_latent.pdf}\label{fig:os_latent}
% \caption{Second.}%
% \label{fig:2figsB}%
% \end{minipage}%
\subfigure[normal][latents of osillator]{\includegraphics[width=0.2\linewidth]{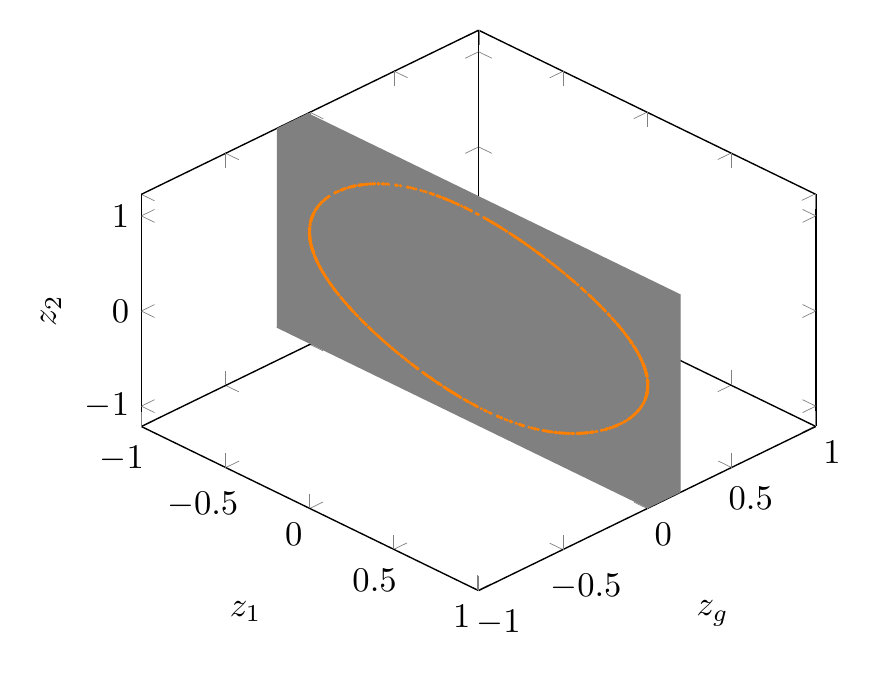}\label{fig:os_latent}}
\subfigure[comparison between latent and $x_1$]{\includegraphics[width=0.2\linewidth]{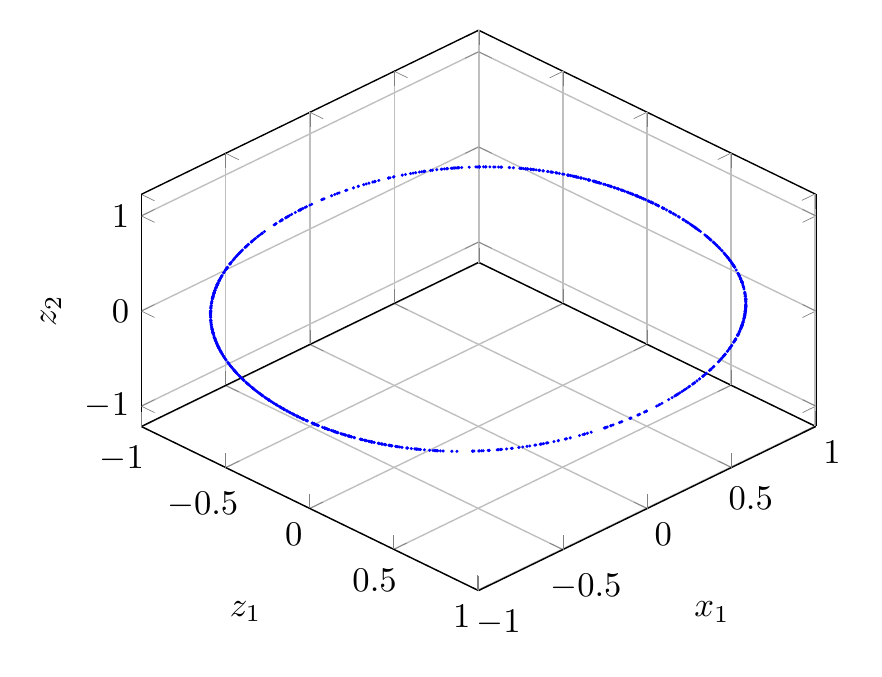}}
\subfigure[comparison between latent and $v$]{\includegraphics[width=0.2\linewidth]{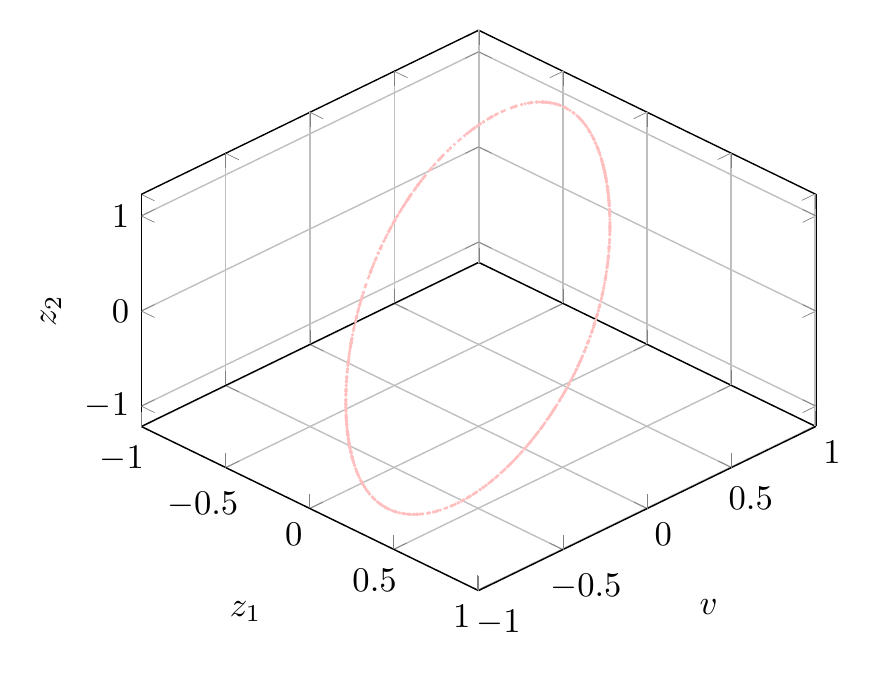}\label{fig:os_p}}
\subfigure[latent of the ball's orbit]{\includegraphics[width=0.2\linewidth]{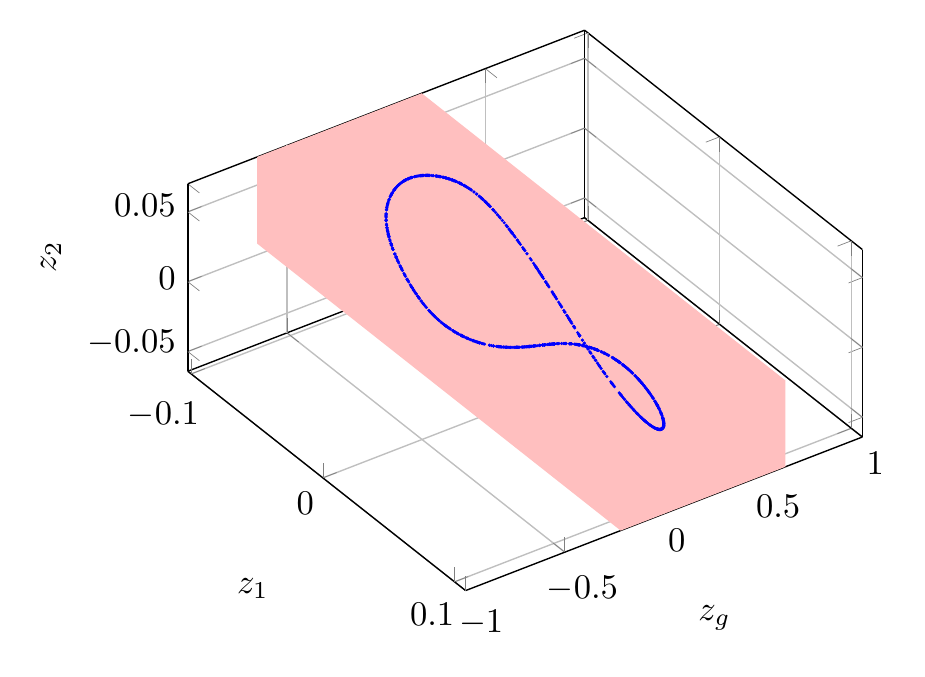}\label{fig:orbit}}\\
\subfigure[comparison between latent and $x$]{\includegraphics[width=0.2\linewidth]{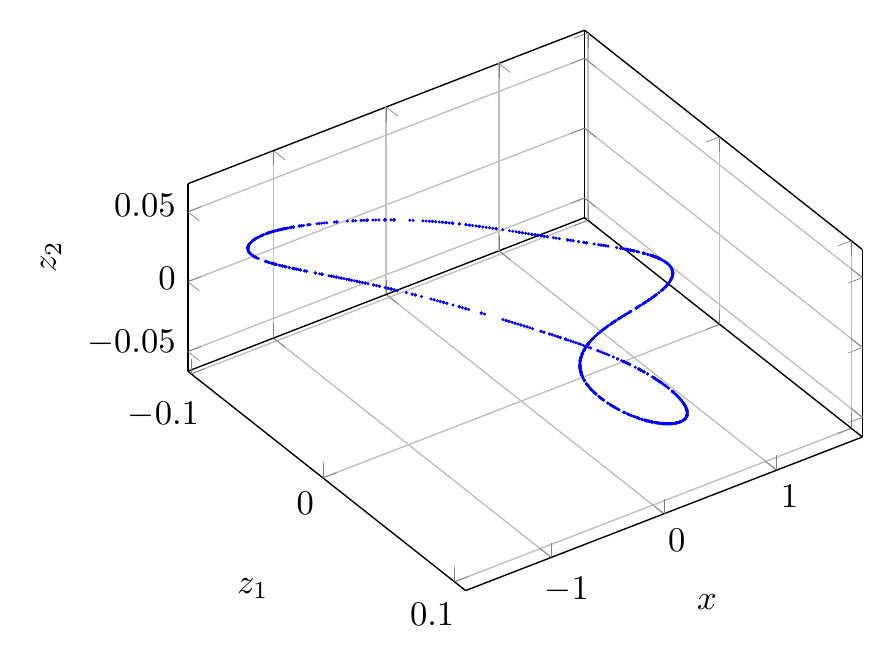}\label{fig:orbit_x}}
\subfigure[comparison between latent and $y$]{\includegraphics[width=0.2\linewidth]{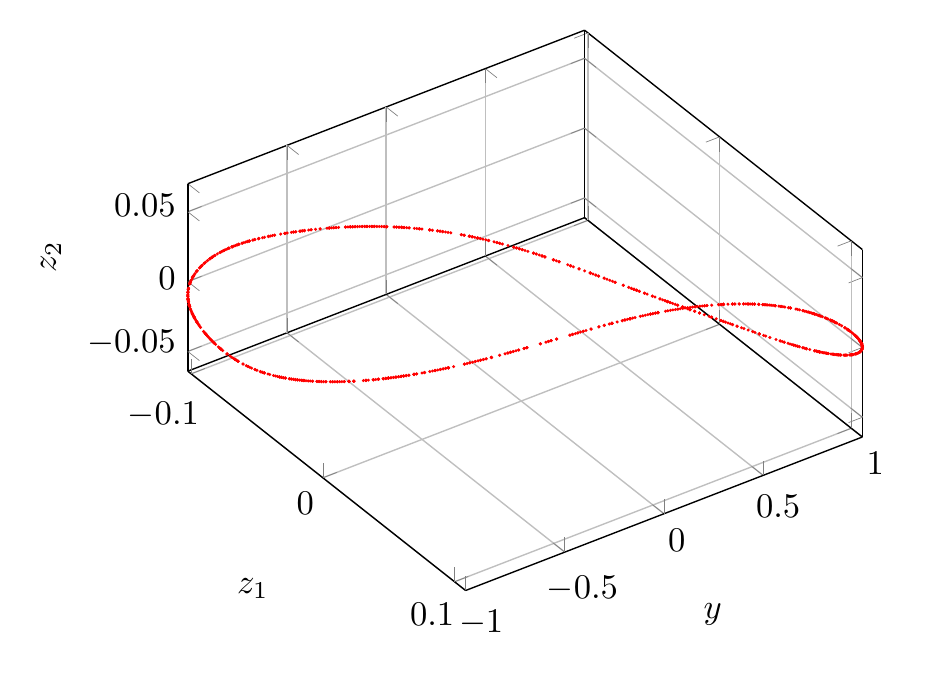}\label{fig:orbit_y}}
\subfigure[latent of the observations of quantum states]{\includegraphics[width=0.2\linewidth]{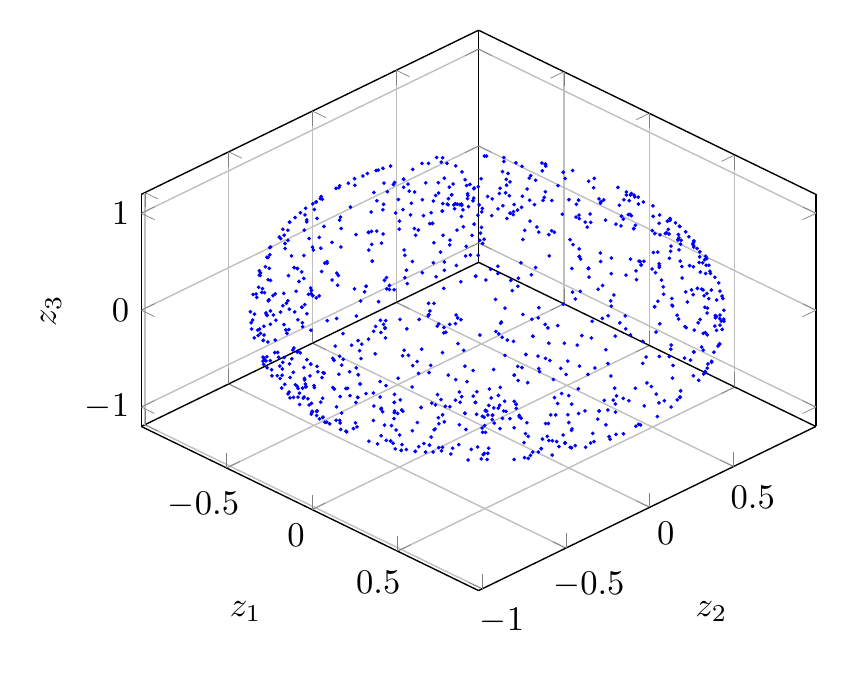}\label{fig:state}}
\subfigure[comparison between the azimuths]{\includegraphics[width=0.2\linewidth]{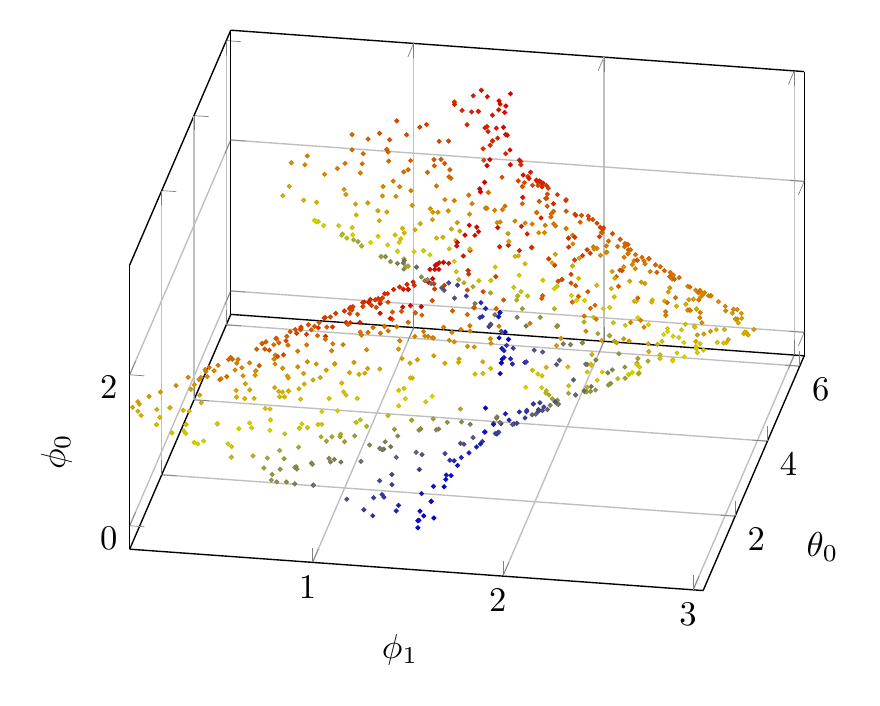}\label{fig:state_theta}}
\label{fig:total}
\caption{
Fig ~\ref{fig:os_latent} to ~\ref{fig:os_p} belong to the first numerical simulation. \ref{fig:orbit} to \ref{fig:orbit_y} belong to the second numerical simulation. \ref{fig:state} and \ref{fig:state_theta} belong to the third simulation. $z_1$ and $z_2$ denote the TPVs and $z_g$ denotes the GPV.
\ref{fig:os_latent} and \ref{fig:orbit} shows that the GPV is always near zero, which means that we don't need a GPV in these model. For the case of quantum state, we compare the polar coordinates of a quantum state and the azimuths of effective Bloch sphere. 
}
\end{figure*}

We test this neuron network by numerical simulation. Three toy models whose manifolds are different are considered here. % All parameters used in the numerical simulation are listed in Appendix ~\ref{}.

We use pytorch to implement neural networks. Our code can be found here. We use the same structure in the encoding and decoding network. They have two hidden layers, each with 20 neurons. In encoder network, we choose Tanh as the active function, while in decoder network, we choose ReLU as the active function.
In numerical simulation, we first generate a database of $1000\times m$, where m is the dimension of the observations. We then use the TDA tools to analyze the Betti numbers of the manifold constituted by observations and set up the latent neurons. We used the Adam optimizer and set the learning rate to $0.0001$. At each training session, we randomly select 100 sets of data from the database as a batch, then calculate the average of its loss and reverse propagate it. All training can be done on the desktop in less than 6 hours.

\subsection{classical Harmonic oscillator}

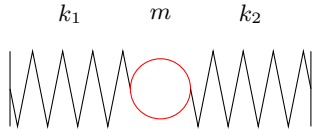
\begin{figure}
    \begin{tikzpicture}
    \draw [-] (0,0) --(0,1);
    \draw [-] (0,0.5)--(0.1,0)--(0.3,1)--(0.5,0)--(0.7,1)--(0.9,0)--(1.1,1)--(1.3,0)--(1.5,1)--(1.6,0.5);
    \draw [red] (2,0.5) circle [radius = 0.4];
    \draw [-] (2.4,0.5)--(2.5,0)--(2.7,1)--(2.9,0)--(3.1,1)--(3.3,0)--(3.5,1)--(3.7,0)--(3.9,1)--(4,0.5);
    \draw [-] (4,0)--(4,1);
    \node at  (0.8,1.5) {$k_1$};
    \node at  (2,1.5) {$m$};
    \node at  (3.2,1.5) {$k_2$};
    \end{tikzpicture}
    \caption{The conversed system consists of two spring connected to a ball. The spring constants $k_1$,$k_2$ and the mass of the ball keep unchanged during the experiment. The total energy comprises of the potential energy of two springs and the kinetic of the ball.}
    \label{harmonic}
\end{figure}

One type of vital importance is the system with some conserved quantity. In classical mechanics, we study these systems by their phase space, constituted by all possible states. Generalized coordinates and generalized momentum are usually not independent of each other, but it is not possible to uniquely determine the other when only one is known. This is caused by the topological nature of the phase space. In some cases, the experimenter may only be able to make observations and cannot exert influence on the observation object. At this time, the observations may constitute a compact manifold, and the traditional VAE network cannot accurately reduce the dimension of the parameter space.

The most common conversed system is the harmonic oscillator, which is a conservative system. We first consider a ball connected with two springs, as shown in Fig~\ref{harmonic}. We can write the energy as 
\begin{equation}
    E = \frac{1}{2} m v^2 + \frac{1}{2} k_1 (x-\frac{1}{2})^2 + \frac{1}{2} k_2 (x+\frac{1}{2})^2
\end{equation}
Here we make $m=1$ and $k_1 = k_2 =1$. The unit is not important. In loss function ~\ref{loss} we make $[\alpha,\beta,\gamma] = [1,1,100]$. In this system, the underlying changing physical variables are $\mathcal{P}^{(k)} = (x_1,v)$ or $(x_2,v)$. The observations one can choose here are $\mathcal{E}^k = \{x_1,x_2,v\}$ where $x_1$ and $x_2$ denote the distance from the ball to the bottom of two springs, respectively, and $v$ denote the speed of the ball. We specify that the direction of speed is positive to the right and negative to the left. We generate the observations $\mathcal{E}^{(k)}$ by randomly sampling from the evolution of the system. According to classical mechanics, we know that the manifold should be a circle embedded in 3-d Euclid space. Programme shows that the Betti numbers are $(1,1,0)$. It means that we need 2 latent variables $(z_1,z_2)$ and the topological term in loss function ~\eqref{loss} is $\mathcal{T} = |z_1^2 + z_2^2 -1|$. Besides, we set up one general physical variable, whose prior $P(z_g)$ is a Gaussian distribution.

After the train is finished, we calculate the latent variables corresponding to the observations in $\mathcal{E}$ by encoder network. As is shown in Fig~\ref{fig:os_latent}, the latent variables constitute a circle again. Fig ~\ref{fig:os_latent} shows that the GPV $z_g$ is always zero for different observations. This means that the effective physical variables are $\Tilde{\mathcal{P}^{(k)}} = (z_1,z_2)$.

We compared the new physical variables $(z_1,z_2)$ with observations $(x_1,v)$, as shown in Fig~\ref{fig:os_x} and ~\ref{fig:os_p}. Given effective physical variables $\Tilde{P}$, we can uniquely determine $P$ while the mapping is continuous. They are one-to-one correspondence, so $\Tilde{P}$ can be used to build a theory.
\subsection{Orbit}

\begin{figure}
    \begin{tikzpicture}
    \draw[->] (-2,0,0) --(2,0,0);
    \draw[->] (0,-2,0) --(0,2,0);
    \draw[->] (0,0,-3) --(0,0,3);
    \draw[red] (0,0,3) circle [radius=0.4] ;
    \node at (-1,0,3) {ball-2};
    \draw[blue] (2,0,0) circle [radius=0.4];
    \node at (2,-0.6,0) {ball-1};
    \draw[orange,->] (0,-0.7,2) arc (270:340:1 and 2);
    \draw[orange,->] (2,0.7,0) arc (0:90:1);
    \node at (2,1.7,0) {$\omega$};
    \node at (0.5,-0.7,2) {$2\omega$};
    % \filldraw [black] (0,0) circle [radius = 0.1];
    % \node at (1,1.8) {ball-1};
    % \node at (-2,-2.3) {ball-2};
    % \draw [red] (0,0,1) circle [radius = 0.4];
    % \draw [blue] (1,1) circle [radius = 0.4];
    % \draw [->] (1.3,0.5) .. controls (1.4,0) .. (1.0,-0.5);
    % \node at (1.5,0) {$\omega$};
    % \draw [->] (-2,-1) .. controls (-2,-0.5) .. (-1.5,0);
    % \node at (-2.2,-0.8) {$2\omega$};
    \end{tikzpicture}
    \caption{Two small balls rotate around the same immobility point. The two balls move at a constant speed circumference on the planes of the two meetings. One ball has twice the angular speed of the other.}
    \label{orbit}
\end{figure}
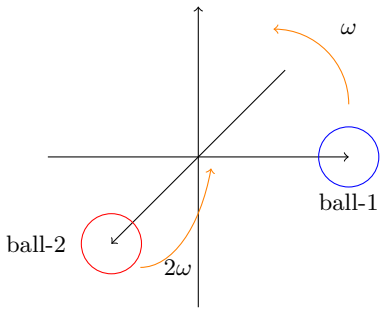

In classical physics and special relativity, the inertial reference system plays an important role. The physical laws in an inertial reference frame are usually simpler than those in a non-inertial reference frame. In fact, there was no concept of an inertial reference frame at the beginning. Due to the existence of gravity, there is no truly perfect inertial reference frame. When observing some simple motions in a non-inertial frame, the observations may constitute some complex manifolds. 

%Another case where the manifold has holes is the trajectory of motion in a noninertial reference system. We can also find some laws in the noninertial reference system, while the law is more complicated. 

Consider a scenario where two balls, labeled by $1$ and $2$, rotate around a fixed point, as shown in Fig~\ref{orbit}. Both balls do a constant-speed circular motion. We assume that the distance between it and the fixed point is $1$. Establishing the Cartesian coordinate system with a fixed point as the origin, $ball-1$ starts at $(1,0,0)$ and $ball-2$ starts at $(0,0,1)$.They have the same radius of rotation, but they are on different planes. $ball-1$ moves in the $z=0$ plane and $ball-2$ moves in the $y=0$ plane. $ball-1$ has the angular speed of $\omega_1 = 1$ and $ball-2$ has the angular speed of $\omega_2 = 2$. Unit is not important. Observations $\mathcal{E}^k = (x,y,z)$ is the three-dimensional coordinates of $ball-1$ measured by $ball-2$ as the reference system.
%They have the same radius of rotation, but they are on different planes. They have different Angular velocities. Observations $\mathcal{E}^k = (x,y,z)$ is the three-dimensional coordinates of the other ball measured by one of the balls as the reference system. With the parameters listed in Appendix ~\ref{}, $\mathcal{E}$ form a manifold with two holes.

We generate the observations $\mathcal{E}^{(k)}$ by random sampling from the trajectory With TDA we know the betti numbers of $\mathcal{E}$ are $(1,2,0)$ so the topological term is $|(z_1^2+z_2^2)^2 - 0.01 \times (z_1^2 - z_2^2)|$, and again We set up one GPV. The hyperparameters of loss function $\ref{loss}$ is $[\alpha,\beta,\gamma] = [1,100,100]$.

After training, the latent variables corresponding to the observations $\mathcal{E}$ are plotted, as shown in fig ~\ref{fig:orbit}. The GPV is always zero for different observations, which means the effective physical variables is $\Tilde{P} = (z_1,z_2)$, like the first example. Fig ~\ref{fig:orbit_x} and ~\ref{fig:orbit_y} shows the comparison between the effective physical variables and observations. The results show that $\Tilde{P}$ and observational measurements are still one-to-one, so $\Tilde{P}$ is an effective representation.

\subsection{Quantum state}
Both of the previously introduced situations come from the limitations of experimental conditions. If we can modify the conserved quantities of conservative systems through experiments, or find some approximate inertial reference frames, such as the earth or distant stars, then it is possible to turn it into a situation that can be solved by traditional VAE networks. Unlike these, the topological properties of some observations are derived from the laws of physics. If this physical quantity can only take partial values in a certain experimental system, then the VAE network may work at this time. But if we want to establish a physical theory, you need to have a comprehensive understanding of the key parameters, and then you need a VAE network based on topological properties.

In quantum mechanics, quantum state is described by wave function, i.e. a vector in Hilbert space. Here we consider a two-level system, which according to quantum mechanics can be described as $\ket{\psi} = a\ket{0} + b\ket{1}$, where $|a|^2  + |b|^2 = 1$. The machine don't know how to describe this system, but it will learn some efficient variables. Suppose we can get five observations in experiments,i.e. $\mathcal{E}^k = \{\mathcal{O}^1,\mathcal{O}^2,\dots,\mathcal{O}^5 \}$. In experiments, $\mathcal{E}$ is due to the experimental setup. In our numerical simulation, we calculate observations by $\mathcal{O}^i =  \bra{\psi} O^i \ket{\psi} $, where $O^i$ is the pauli matrix and the combination of pauli matrix:
\begin{equation}
\begin{split}
    \mathcal{O}_1 = \left[
    \begin{array}{cc}
    0     & 1 \\
    1     & 0
    \end{array}
    \right] ,
   \mathcal{O}_2 = \left[
    \begin{array}{cc}
    0     & i \\
    -i     & 0
    \end{array}
    \right],
       \mathcal{O}_3 = \left[
    \begin{array}{cc}
    1     & 0 \\
    0     & -1
    \end{array}
    \right]\\
       \mathcal{O}_4 = \left[
    \begin{array}{cc}
    0     & 1+i \\
    1-i     & 0
    \end{array}
    \right],
       \mathcal{O}_5 = \left[
    \begin{array}{cc}
    1     & i \\
    -i     & -1
    \end{array}
    \right]
\end{split}
\end{equation}

We need many different states, so the coefficient $a$ and $b$ of the wave function are the physical variables that change as the experiment progresses. In this model, $\mathcal{P}$ consists of wave functions we use and $\mathcal{P}^k = \{a^k , b^k\}$. The first step is to characterize the topological feature of $\mathcal{E}$. By TDA, we can find that the Betti numbers of the data set are $[1,0,1]$, which means that the manifold of $\mathcal{E}$ is homeomorphic to a sphere. This can be understood because we can use the point on the Bloch sphere to represent the state of a two-level system. So that we know the number of TVs is $3$. However, we don't know how many GVs are. Different from TVs, we can assume GVs are independent of each other. In general, we do this by assuming $P(\Vec{z})$ in \eqref{loss} is independent Gaussian distribution. In this case, we only introduce one GV.

As shown in Fig ~\ref{fig:state}, after training, TVs form a sphere. At the same time GV stabilizes near 0(not drawn in the figure). That means the  five observations can be reduced to three variables continuously. We call these three variables as equivalent density matrix, denoted by $\Tilde{\rho}$. Only two of them are independent. We can construct two independent angles $(\theta,\phi)$ by transforming Cartesian coordinates to polar coordinates. In figure ~\ref{fig:state_theta} we compare $(\theta_0,\phi_0)$ and $\phi_1$ corresponding to the equivalent density matrix. We can see that these two representations have a one-to-one correspondence. Unlike before, the mapping here is discontinuous.
\section{Conclusion}
In this work, we discuss the defect of the traditional VAE network and propose a simple solution. We can extract the minimum effective physical variables from the experimental data with the improved method by classifying the latent variables. We test our approach on three models. They represent three different situations that may arise that traditional VAE cannot handle. Some of them come from experimental restrictions, and some come from physical laws. We think the latter situation is more essential. However, in the more complex case, the Betti numbers are not the only useful topological features. Two manifolds may have the same Betti numbers but not Homeomorphic. In such a case, more topological features are needed to design reasonable restrictions on latent variables. When the manifold dimension is higher, it may be hard for TDA to calculate the Betti numbers. One efficient way is to firstly reduce the dimensions by traditional autoencoder, and then calculate the Betti numbers of latent variables. Another important question is how to ensure that the relation between meaningful variables and observations is simple. We leave it for the future work.

\appendix

\section{persistent homology and betti numbers}
Persistent homology is a useful tool for analyzing topological data. The first step is to generate a simplicial complex from the observations. Vietoris–Rips complex is a way of forming a topological space from distances in a set of points. This method uses a generalisation of an $\epsilon$ neighbouring graph
and the final complex is 
\begin{equation}
    \mathcal{V}_{\epsilon} = \{\sigma | r(u,v) < \epsilon, (u,v)\in \sigma\}
\end{equation}
Here $\sigma$ denotes the simplex and $u,v$ are two data points. $r$ is the Euclidean metric. A $k$-simplex $\sigma$ is expressed as $\sigma = [p_1,p_2,\dots,p_{k+1}]$, where $p$ denotes the point in the space. Given a set of $k$-simplex $\sigma_i$, one $k$-chain is defined by 
\begin{equation}
    c = \Sigma_i c_i \sigma_i
\end{equation}
Here the coefficients $c_i$ take values in some field. For the simplicial complex $\mathcal{V}$, we denote the set of all $k$-chains by $C_k(\mathcal{V})$. We can introduce the addition between two $k$-chain. For $c = \Sigma_i c_i \sigma_i$ and $d = \Sigma_i d_i \sigma_i$, the addition is $c+d = \Sigma_i (c_i+d_i)\sigma_i$. Furthermore, the set of all $k$-chains $C_k(\mathcal{V})$ form a abelian group $(C_k(\mathcal{V}),+)$.

For different $k$, a natural group homomorphism is the boundary operator
\begin{equation}
    \partial_k: C_k(V) \rightarrow C_{k-1}(V)
\end{equation}
Boundary operator is linear, and it can be defined by the simplex
\begin{equation}
    \partial_k(\sigma_k) = \Sigma_{i=0}^{i=k}(-1)^i\langle p_0,\dots,p_{i-1},p_{i+1},\dots,p_k \rangle 
\end{equation}
One important subgroup of $C_k(\mathcal{V})$ is the kernel of the boundary operator $Z_k$, namely $k$-cycles. There is also an important subgroups of $Z_k$, denoted by $B_k$, which is the imagine $B_k(V) =Im( \partial_{k+1} C_{k+1}(\mathcal{V}))$.

Both $Z_k$ and $B_k$ are normal, so we can define the quotient group $H_k = Z_k/B_k$, called the $k$-th homology group. The rank of $H_k$ is called the $k$-th Betti number. 

\section{Comparison with other work}
As we point in the paper, the topological feature of latent space will influence the construction of the output. In the field of machine learning, this phenomenon is called manifold mismatch~\cite{de2018topological,davidson2018hyperspherical,rey2019diffusion,gong2019lie}, which will lead to poor representations. 

Let's consider what happens when manifold mismatch occurs. Recall the first example, suppose we have a latent structure as $S^1$, i.e., a circle. While in normal VAE, the prior distribution $p(z)$ will limit the latent structure to be $R^n$, where $n$ is the number of latent variables. This prior will make the hidden variables independent of each other. Like before, here we set $n=3$. One parameter can describe this system, e.g. the angle $\theta$. Thus we get a network which maps the coordinate $(x,y)$ to a parameter $\theta$ and then maps the $\theta$ to the coordinate $(x,y)$. In practice, our data set is finite, and the finally network will only be effective for an arc on the circle, instead of the total circle. This problem can be worse when the manifold dimension is higher.
% If we want a network that is effective for the total circle, the encoder must be discontinuous.

In the earlier works, some methods have been developed for capturing the topological features of date set. However, to our understanding, these methods are not suitable for our goal because they usually derive a set of entangled latent vatiables.

\bibliography{ref.bib}
\end{document}